\documentclass[10pt,twocolumn,letterpaper]{article}

\usepackage{wacv}
\usepackage{times}
\usepackage{epsfig}
\usepackage{graphicx}
\usepackage{amsmath}
\usepackage{amssymb}
\usepackage[position=bottom]{subfig}
\usepackage{floatrow}

\usepackage[pagebackref=true,breaklinks=true,letterpaper=true,colorlinks,bookmarks=false]{hyperref}

\wacvfinalcopy 

\begin{document}

\title{\vspace{-1.3cm}ICface: Interpretable and Controllable Face Reenactment Using GANs }

\author{Soumya Tripathy \\
Tampere University\\
{\tt\small soumya.tripathy@tuni.fi}
\and
Juho Kannala \\
Aalto University of Technology\\
{
\tt\small juho.kannala@aalto.fi}
\and
Esa Rahtu \\
Tampere University\\
{
\tt\small esa.rahtu@tuni.fi}
}
\twocolumn[{%
\renewcommand\twocolumn[1][]{#1}%
\maketitle
\begin{center}
    \centering
    \vspace{-20pt}
    \includegraphics[width=0.9\textwidth,height=7.5cm]{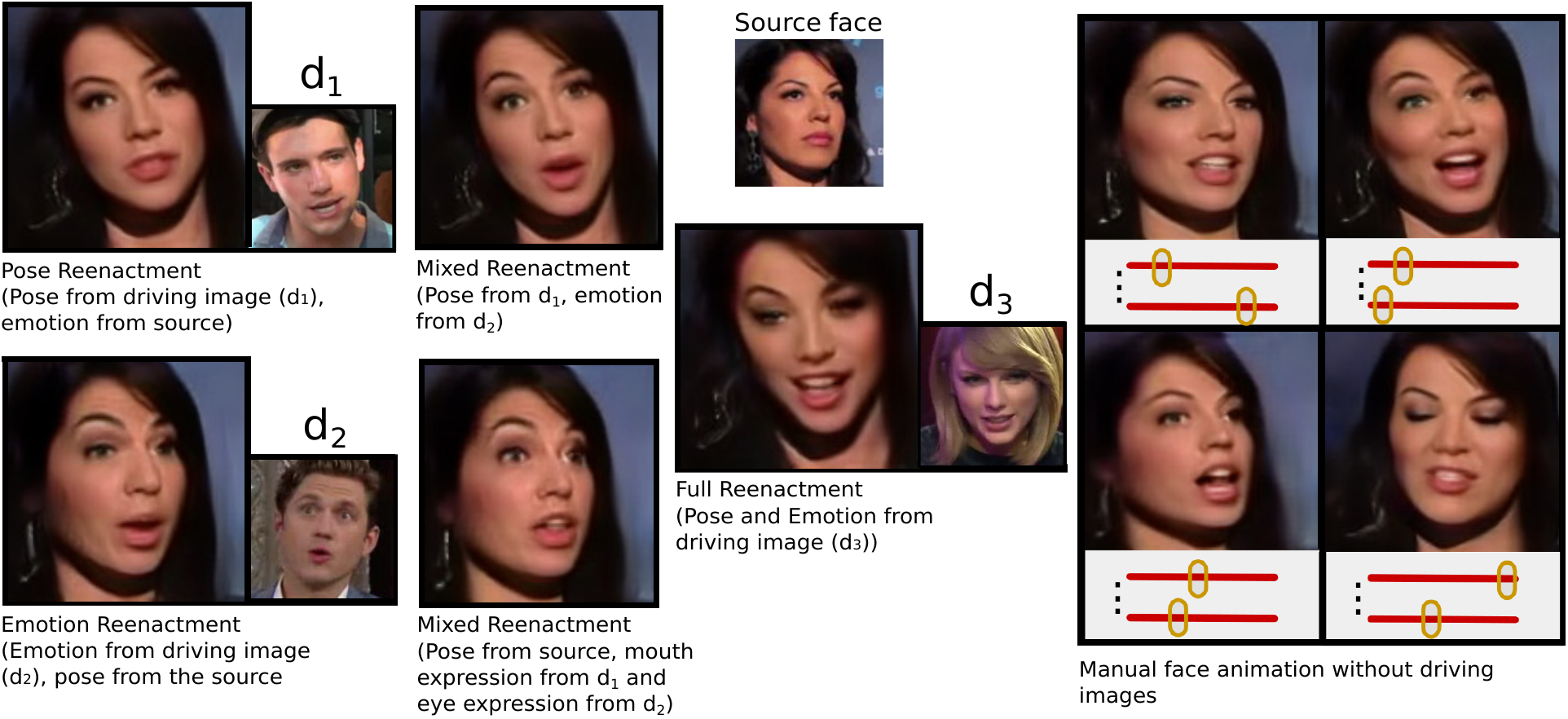}
    \vspace{-5pt}
    \captionof{figure}{Given a single \emph{source} face image and a set of \emph{driving} attributes, our model is able to generate a high quality facial animation. The driving attributes can be selectively specified by one or more driving face images or controlled via interpretable continuous parameters.}
    \label{teaser}
\end{center}%
}]



\begin{abstract}
\vspace{-1.5cm}
This paper presents a generic face animator that is able to control the pose and expressions of a given face image. The animation is driven by human interpretable control signals consisting of head pose angles and the Action Unit (AU) values. The control information can be obtained from multiple sources including external driving videos and manual controls. Due to the interpretable nature of the driving signal, one can easily mix the information between multiple sources (e.g.~pose from one image and expression from another) and apply selective post-production editing. The proposed face animator is implemented as a two stage neural network model that is learned in self-supervised manner using a large video collection. The proposed \textbf{I}nterpretable and \textbf{C}ontrollable \textbf{face} reenactment network (ICface) is compared to the state-of-the-art neural network based face animation techniques in multiple tasks. The results indicate that ICface produces better visual quality, while being more versatile than most of the comparison methods. The introduced model could provide a lightweight and easy to use tool for multitude of advanced image and video editing tasks. The program code will be publicly available upon the acceptance of the paper.
\end{abstract}
\vspace{-0.5cm}
\section{Introduction}
The ability to create a realistic animated video from a single face image is a challenging task. It involves both rotating the face in 3D space as well as synthesising detailed deformations caused by the changes in the facial expression. A lightweight and easy-to-use tool for this type of manipulation task would have numerous applications in animation industry, movie post-production, virtual reality, photography technology, video editing and interactive system design, among others.

Several recent works have proposed automated face manipulation techniques. A commonly used procedure is to take a \emph{source} face and a set of desired facial attributes (e.g. pose) as an input and produce a face image depicting the source identity with the desired attributes. The source face is usually specified by one or more example images depicting the selected person. The facial attributes could be presented by categorical variables, continuous parameters or by another face image (referred as a \emph{driving} image) with desired pose and expression. 

Traditionally, face manipulation systems fit a detailed 3D face model on the source image(s) that is later used to render the manipulated outputs. If the animation is driven by another face image, it must also be modelled to extract the necessary control parameters. Although these methods have reported impressive results (see e.g.~Face2Face \cite{c33},\cite{c37}), they require complex 3D face models and considerable efforts to capture all the subtle movements in the face. 

Recent works \cite{x2face,c15} have studied the possibility to bypass the explicit 3D model fitting. Instead, the animation is directly formulated as an end-to-end learning problem, where the necessary model is obtained implicitly using a large data collection. Unfortunately, such implicit model usually lacks interpretability and does not easily allow selective editing or combining driving information from multiple sources. For example, it is not possible to generate an image which has all other attributes from the driving face, except for an extra smile on the face. Another challenge is to obtain expression and pose representation that is independent of the driving face identity. Such disentanglement problem is difficult to solve in a fully unsupervised setup and therefore we often see that the identity specific information of the driving face is "leaking" to the generated output. This may limit the relevant use cases to a few identities or to faces with comparable size and shape. 

In this paper, we propose a generative adversarial network (GAN) based system that is able to reenact realistic emotions and head poses for a wide range of source and driving identites. Our approach allows further selective editing of the attributes (e.g.~rotating the head, closing the eyelid etc.) to produce novel face movements which were not seen in the original driving face. The proposed method offers extensive human interpretable control for obtaining more versatile and high quality face animation than with the previous approaches. Figure \ref{teaser} depicts a set of example results generated by manipulating a single source image with different mixtures of driving information. 

The proposed face manipulation process consists of two stages: 1) extracting the facial attributes (emotion and pose) from the given driving image, and 2) transferring the obtained attributes to the source image for producing a photorealistic animation. We implement the first step by representing the emotions and facial movements in terms of Action Units (AUs) \cite{c17} and head pose angles (pitch, yaw and roll). The AU activations \cite{c17} aim at modelling the specific muscle activities and each combination of them can produce different facial expression \cite{c17,gani}. Our main motivation is that such attributes are relatively straightforward to extract from any facial image using publicly available software and this representation is fairly independent of the identity specific characteristics of the face.

We formulate the second stage of the face animation process using a conditional generative model on the given source image and facial attribute vector. In order to eliminate the current expression of the source face, we first map the input image to a neutral state representing frontal pose and zero AU values. Afterwards, the neutral image is mapped to the final output depicting the desired combination of driving attributes (e.g.~obtained from driving faces or defined manually). As a result, we obtain a model called \textbf{I}nterpretable and \textbf{C}ontrollable \textbf{face} reenactment network (ICface). 

We make the following three contributions. i) We propose a data driven and GAN based face animation system that is applicable to a large number of source and driving identities. ii) The proposed system is driven by human interpretable control signals obtainable from multiple sources such as external driving videos and manual controls. iii) We demonstrate our system in multiple tasks including face reenactment, facial emotion synthesis, and multi-view image generation from single-view input. The presented results outperform several recent (possibly purpose-built) state-of-the-art works. 

\begin{figure}
\vspace{-0.3cm}
\begin{center}
\includegraphics[width=8.2cm]{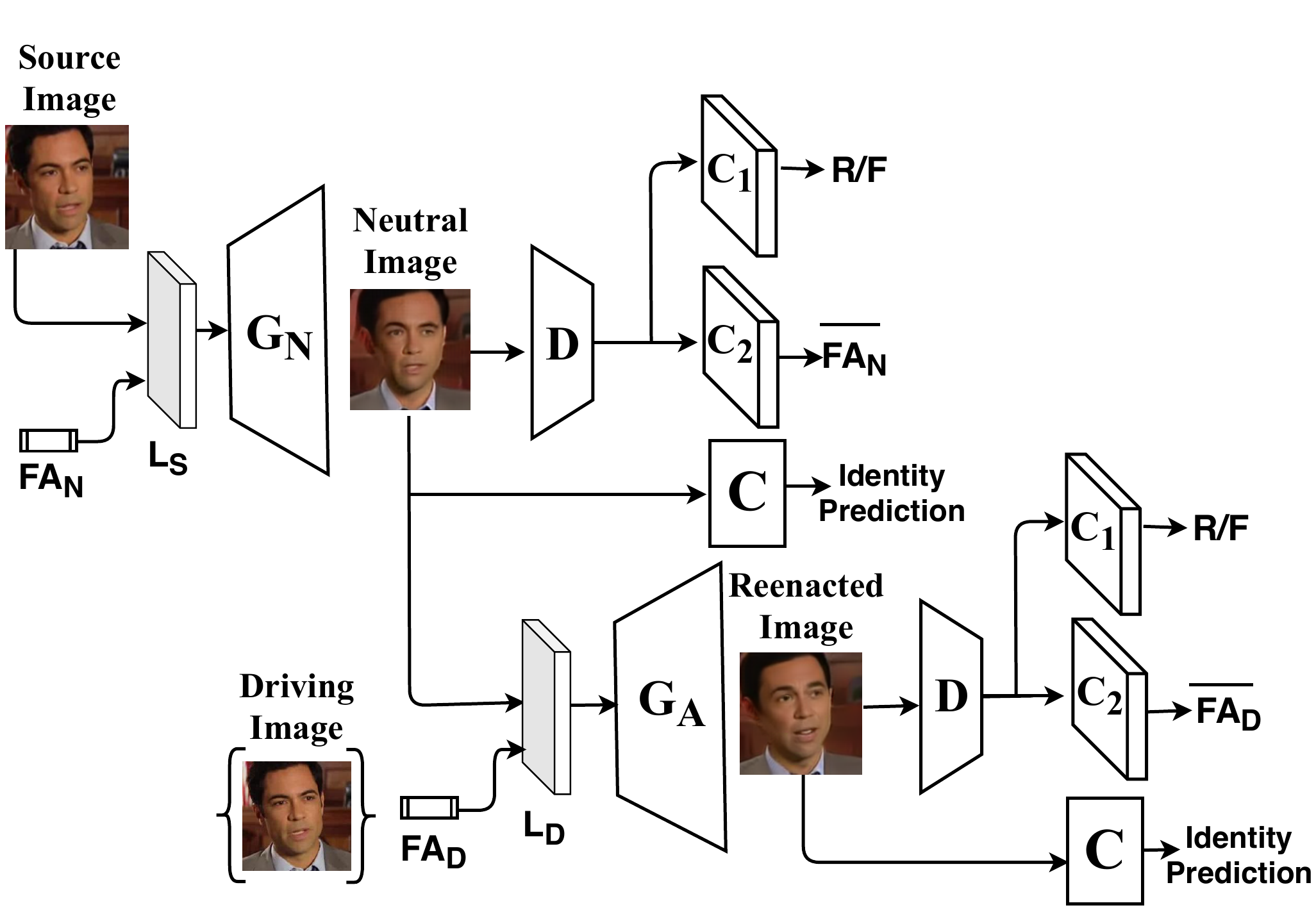}
\end{center}
\vspace*{-8pt}
   \caption{The overall architecture of the proposed model (ICface). During the training, two frames are selected from the same video and denoted as source and driving images. The generator \(G_N\) takes the source image and neutral facial attributes ($FA_N$) as input and produces the source identity with central pose and neutral expression (neutral image). In the second phase, the generator \(G_A\) takes the neutral image and attributes extracted from the driving image ($FA_D$) as an input and produces an image with the source identity and driving image's attributes. The generators are trained using multiple loss functions implemented using the discriminator D (see Section 3 for details). In addition, since the driving and source images have the same identity, a direct pixel based reconstruction loss can also be utilized. Note that this is assumed to be true only during training and in the test case the identities are likely to be different.\vspace{-0.6cm}}
\label{archi}
\end{figure}

\vspace{-0.8cm}
\section{Related work}
\vspace{-0.2cm}
The proposed approach is mainly related to face manipulation methods using deep neural networks and adversarial generative networks. Therefore, we concentrate on reviewing the most relevant literature under this scope.
\subsection{Face Manipulation by Generative Networks} 

Deep neural networks are very popular tools for controlling head pose, facial expressions, eye gaze, etc. Many works \cite{c2,c3,c4,c5} approach the problem using supervised paradigm that requires plenty of annotated training samples. While such data is expensive to obtain, the recent literature proposes several unsupervised and self-supervised alternatives \cite{c1, c7, c8, c6}. 

In \cite{c9}, the face editing was approached by decomposing the face into a texture image and a deformation field. After decomposition, the deformation field could be manipulated to obtain desired facial expression, pose, etc. However, this is a difficult task, partially because the field is highly dependent on the identity of the person. Therefore, it would be hard to transfer attributes from another face image.


Finally, X2Face \cite{x2face} proposes a generalized facial reenactor that is able to control the source face using driving video, audio, or pose parameters. The transferred facial features were automatically learned from the training data and thus lack clear visual interpretation (e.g. close eyes or smile). The approach may also leak some identity specific information from the driving frames to the output. X2Face seems to work best if the driving and source images are from the same person. 


\subsection{Face Manipulation with GANs} 

The conditional variant of the Generative Adversarial Network (GAN) \cite{c19,c18} have received plenty of attention in image to image domain translation with paired data \cite{c20}, unpaired data \cite{c21}, or even both \cite{c22}. Similar GAN based approaches are widely used for facial attribute manipulation in many supervised and unsupervised settings \cite{c10,c11,c12,c13}. The most common approach is to condition the generator on discrete attributes such as blond or black hair, happy or angry, glasses or no glasses and so on. A recent work \cite{gani} proposed a method called GANimation that was capable of generating a wide range of continuous emotions and expressions for a given face image. They utilized the well known concept of action units (AUs) as a conditioning vector and obtained very appealing results. Similar results are achieved on portrait face images in \cite{c31,c32}. Unfortunately, unlike our method, these approaches are not suitable for full reenactment, where the head pose has to be modified. Moreover, they add the new attribute directly on to the existing expression of the source image which can be problematic to handle. 

In addition to the facial expression manipulation, the GANs are applied to the full face reenactment task. For instance, the CycleGAN \cite{c14} could be utilised to transform expressions and pose between image pair (see examples in \cite{x2face}). Similarly, the ReenactGAN \cite{c15} is able to perform face reenactment, but only for a limited number of source identities. In contrast, our GAN based approach generalizes to a large number of source identities and driving videos. Furthermore, our method is based on interpretable and continuous attributes that are extracted from the driving video. The flexible interface allows user to easily mix the attributes from multiple driving videos 
and edit them manually at will. 
In the extreme case, the attributes could be defined without any driving video at all.

\section{Method}
\vspace{-0.2cm}
The goal of our method is to animate a given \emph{source} face in accordance to the facial attribute vector \((FA_D =\left[ \textbf{p}^T, \textbf{a}^T \right]^T)\) that consists of the head pose parameters \(\textbf{p}\)  and action unit (AU) activations \(\textbf{a}\). More specifically, the head pose is determined by three angles (pitch, yaw, and roll) and the AUs represent the activations of 17 facial muscles \cite{c17}. In total, the attribute vector consists of 20 values determining the pose and the expression of a face. In the following, we will briefly outline the workflow of our method. The subsequent sections and Figure \ref{archi} provide further details of the architecture and training procedure. The specific implementation details are found in the supplementary material. 

In the first stage, we concatenated the input image (size WxHx3) with the neutral facial parameters $FA_N =[\textbf{p}_N,\textbf{0}]$, where $\textbf{p}_N$ refers to the central pose. This is done by first spatially replicating the attribute vector and then channel-wise concatenating the replicated attributes (WxHx20) with the input image. The resulting representation \(L_S\) (WxHx23) is subsequently fed to the neutralisation network \(G_N\) that aims at producing a frontal face image (WxHx3) depicting the source identity with neutral facial expression. 

In the second stage, we concatenated the obtained neutral (source) face image with the \emph{driving} attribute vector \(FA_D\) that determines the desired output pose and AU values. The concatenation is done in similar fashion as in the first stage. 
In our experiments, we used OpenFace \cite{c35,c36} to extract the pose and AUs when necessary. The concatenated result \(L_D\) is passed to the generator network \(G_A\) that produces the final animated output (WxHx3) depicting the original \emph{source} identity with the desired facial attributes \(FA_D\).

\subsection{Architecture}

Our model consists of four different sub-networks: Neutraliser, generator, discriminator and identity preserving network. Their structures are briefly explained as follows: 


\vspace{-13pt} 
\paragraph{Neutralizer \((G_N):\)} The neutralizer is a generator network that transforms the input representation \(L_s\) into a canonical face that depicts the same identity as the input and has central pose with neutral expression. The architecture of the \(G_N\) network consists of strided convolution, residual blocks and deconvolution layers. The overall structure is inspired by the generator architecture of CycleGAN \cite{c14}.

\vspace{-13pt}
\paragraph{Generator \((G_A):\)} The generator network transforms input representation \(L_D\) of the neutral face into the final reenacted output image. The output image is expected to depict the \emph{source} identity with pose and expression defined by the \emph{driving} attribute vector $FA_D$. The architecture of the \(G_A\) network is similar to that of \(G_N\).

\vspace{-13pt}
\paragraph{Discriminator \((D):\)} The discriminator network performs two tasks simultaneously: i) it evaluates the realism of the neutral and reenacted images through \(C_1\); ii) it predicts the facial attributes (\(\bar{FA_N} \) and \(\bar{FA_D}\)) through \(C_2\). The blocks \(C_1\) and \(C_2\) consist of convolution block with sigmoid activation. The overall architecture of $D$ is similar to the PatchGANs \cite{c14} consisting of strided convolution and activation layers. The same discriminator with identical weights is used for $G_N$ and $G_A$. 

\vspace{-13pt}
\paragraph{Identity Preserving Network \((C):\)}

We have used a pretrained network called LightCNN \cite{c38} as \(C\) and kept the weights fixed for the whole training process. It provides the identity features for both generated and source faces which are used in the training as identity preserving loss.       


\begin{figure*}[t]
\vspace{-8pt}
\begin{center}
\includegraphics[width=17.5cm, height=12.3cm]{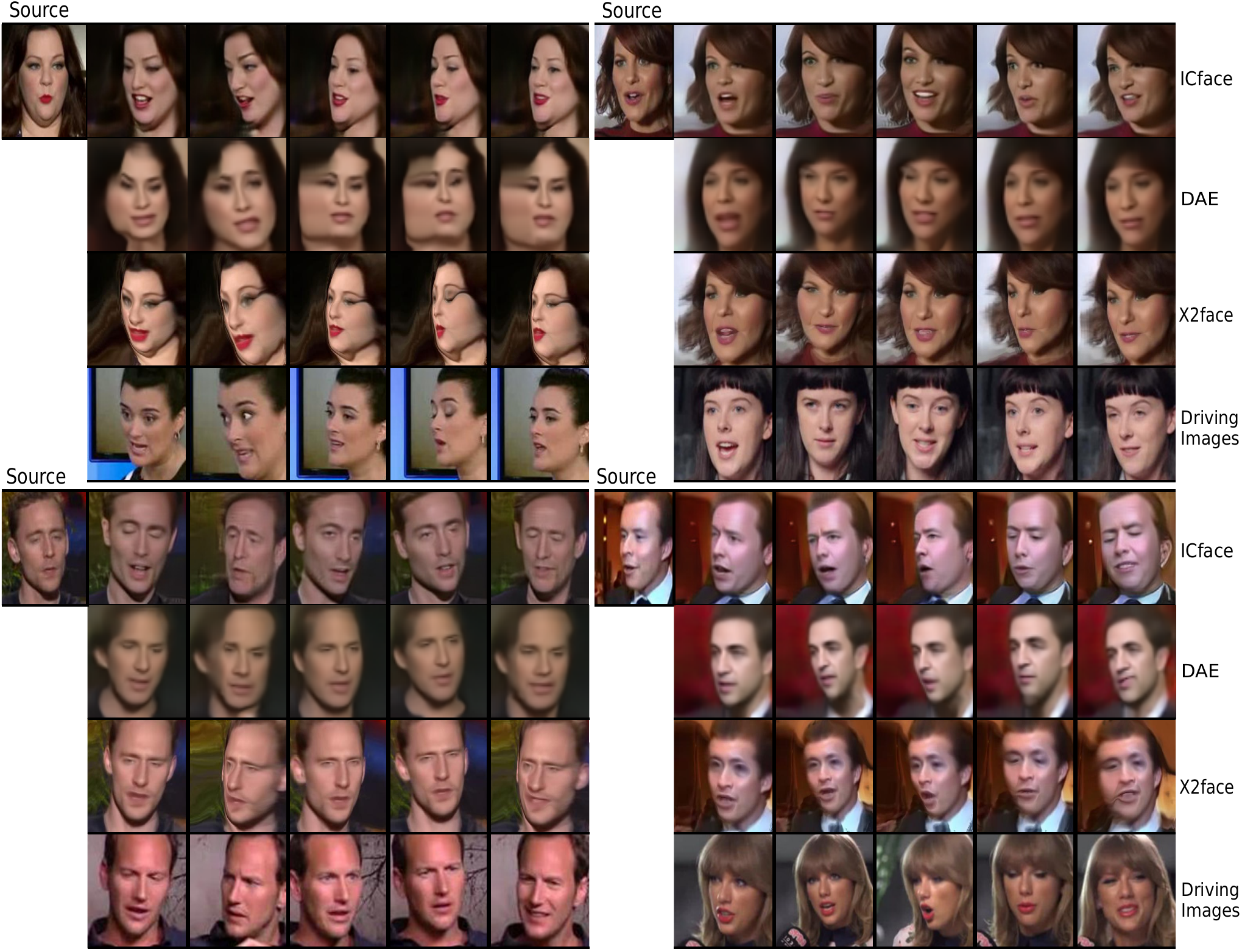} 
\end{center}
\vspace{-3pt}
   \caption{Qualitative results for the face reenactment on VoxCeleb \cite{c23} test set. The images illustrate the reenactment result for four different source identities. For each source, the results correspond to: ICface (first row), DAE \cite{c9} (second row), X2Face \cite{x2face} (third row), and the driving frames (last row). The performance differences are best illustrated in cases with large differences in pose and face shape between source and driving frames.\vspace{-0.3cm}}
\vspace{-0.2cm}
\label{x2face}
\end{figure*}
\subsection{Training the Model}
Following \cite{x2face}, we train our model using VoxCeleb \cite{c23} dataset that contains short clips extracted from interview videos. Furthermore, Nagrani et al.~\cite{c16} provide face detections and face tracks for this dataset, which is utilised in this work. As in \cite{x2face}, we extract two frames from the same face track and feed one of them to our model as a \emph{source} image. Then, we extract the pose and AUs from the second frame and feed them into our model as \emph{driving} attributes $FA_D$. Since both frames originate from the same face track and depict the same identity, the output of our model should be identical to the second frame and it can be treated as a pixel-wise ground truth in the training procedure. All the loss functions are described in the following paragraphs. 

\vspace{-10pt}
\paragraph{Adversarial Loss\((\mathcal{L}_{adv}):\)}

The adversarial loss is a crucial component for obtaining the photorealistic output images. The generator \(G_N\) maps the feature representation \(L_S\) into domain of real images \(X\). Now if \(x \in X\) is a sample from the training set of real images, then the discriminator has to distinguish between \(x\) and \(G_N(L_S)\). The corresponding loss function can be expressed as: 
\begin{equation}
\small
\mathcal{L}_{adv}(G_N,D) =\mathbb{E}_{x}\left[ log D(x)\right] + \mathbb{E}_{L_S}\left[ log( 1-D(G_N(L_S)))\right]
\label{D0n}
\end{equation}
\normalsize 
Similar loss function can also be formulated for \(G_A\) and \(D\) and it would be represented as \(\mathcal{L}_{adv}((G_A,D))\).


\vspace{-10pt}
\paragraph{Facial attribute reconstruction loss\((\mathcal{L}_{FA}):\)}

The generators \(G_N\) and \(G_A\) are aiming at producing photorealistic face images, but they need to do this in accordance with the facial attribute vectors $FA_N$ and $FA_D$, respective. To this end, we extend the discriminator to regress the facial attributes from the generated images and compare them to the given target values. 
The corresponding loss function \(\mathcal{L}_{FA}\) is expressed as: 
\begin{equation}
\small
\begin{aligned}
\mathcal{L}_{FA} =\mathbb{E}_{x}[\Vert C2(D(x))-FA_D \Vert^2_2] + \mathbb{E}_{L_D}[\Vert C2(D(G_A(L_D))) \\ -FA_D\Vert^2_2]
+\mathbb{E}_{L_S}\left[\Vert C2(D(G_N(L_S)))-FA_N\Vert^2_2 \right]
\end{aligned}
\label{D0n}
\end{equation}
\normalsize  
where \(x \in X\) is the driving image with attributes \(FA_D\).
\vspace{-10pt}
\paragraph{Identity classification loss\((\mathcal{L}_{I}):\)}

The goal of our system is to generate an output image retaining the identity of the source person. To encourage this, we have used a pretrained LightCNN \cite{c38} network to comapre the the features of the generated image \((g)\) and source image \((s)\). Specifically we have comapred the features of last pooling layer \((f^p)\) and fully connected layer's \((f^{fc})\) of LghtCNN as follws:
\begin{equation}
\small
\mathcal{L}_{I}=\Vert C(f^p(s))- C(f^p(g)) \Vert_1 + \Vert C(f^{fc}(s))- C(f^{fc}(g)) \Vert_1
\label{D0n}
\end{equation}
\normalsize  
\vspace{-1.0cm}
\paragraph{Reconstruction loss\((\mathcal{L}_{R}):\)}

Due to the specific training procedure described above, we have access to the pixel-wise ground truth of the output. We take advantage of this by applying L1 loss between the output and the ground truth. Furthermore, we stabilize the training of the \(G_N\) by using generated images from \(G_A\) with neutral attributes as a pseudo ground truth. The corresponding loss function is defined as 

\vspace{-15pt}
\begin{equation}
\small
  \begin{aligned}
\mathcal{L}_{R} =\mathbb{E}_{x}[\Vert x - G_A(L_D)\Vert_1] + \mathbb{E}_{L_S}[\Vert G_N(L_S)-G_A(L_S)\Vert_1]
   \end{aligned}
\label{D0n}
\end{equation}
\normalsize 

\vspace{-15pt}
\paragraph{The complete loss function:}

The full objective function of the proposed model is obtained as a weighted combination of the individual loss functions defined above. The corresponding full loss function, with \(\lambda_i\) as regularization parameters, is expressed as 
\vspace{-0pt} 
\begin{equation}
\small
  \begin{aligned}
\mathcal{L} = \mathcal{L}_{adv}(G_N,D) + \mathcal{L}_{adv}(G_A,D) + \lambda_1 \mathcal{L}_{FA} + \lambda_2 \mathcal{L}_{I} + \lambda_3 \mathcal{L}_{R}
   \end{aligned}
\label{final}
\end{equation}
\vspace{-0.95cm}
\section{Experiments}

In all the following experiments, we use a single ICface model that is trained using the publicly available VoxCeleb video dataset \cite{c23}. The video frames are extracted using the preprocessing techniques presented in \cite{c16} and resized to \(128 \times 128\) for further processing. We used 75\% of the data for training and the rest for validation and testing. 
Each component of \(FA\) is normalized to the interval $[0,1]$. The neutral attribute vector \(FA_N\) contains central head pose parameters $[0.5,0.5,0.5]$ and zeros for the AUs. More architectural and training details are provided in the supplementary material.

\begin{figure*}[h]
    \vspace*{-8pt}
    \centering
    \vspace{-0.5\baselineskip}
    \subfloat[Expression Reenactment]{\includegraphics[height=0.57\textwidth]{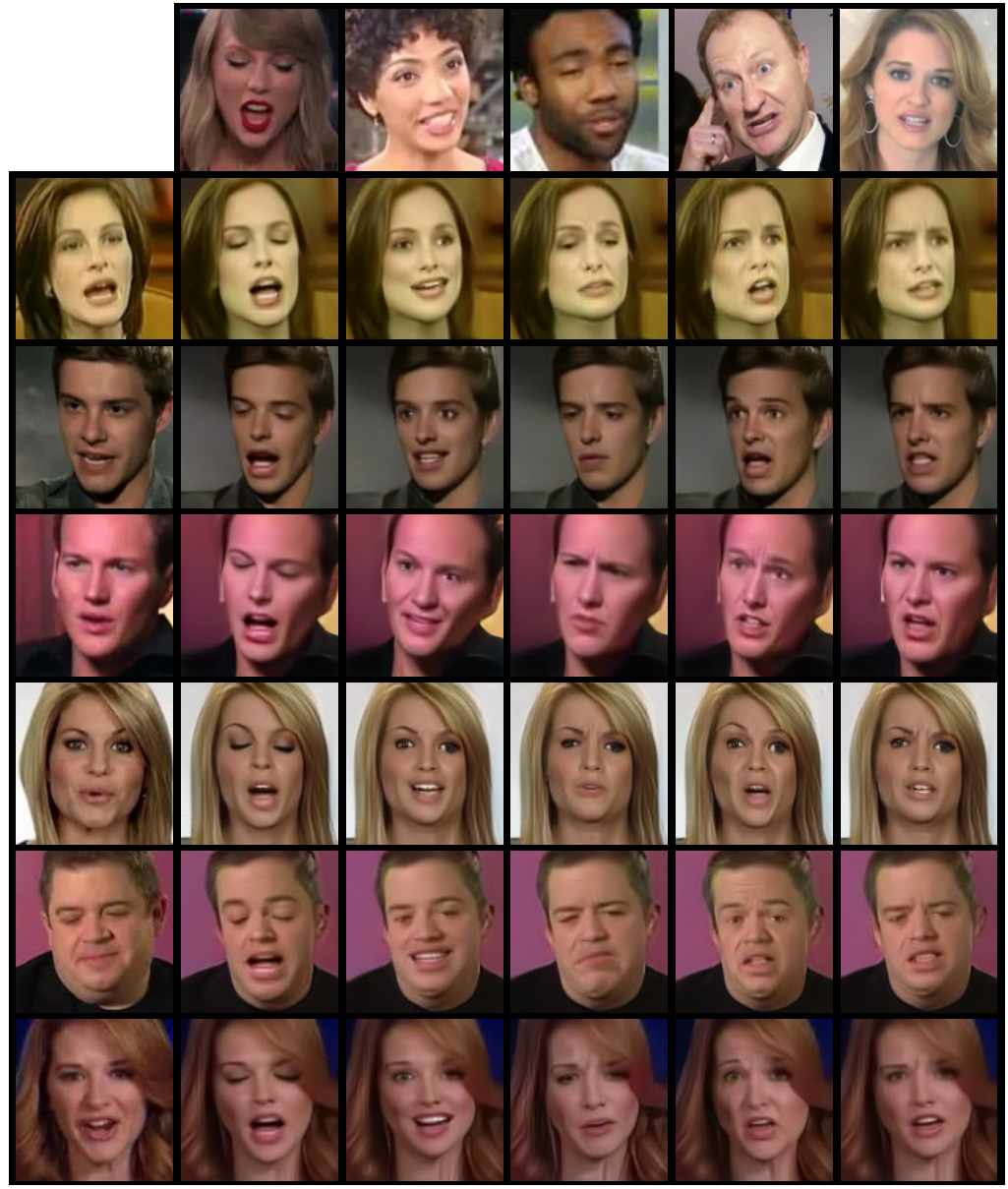}}
    \subfloat[Pose Reenactment]{\includegraphics[height=0.5705\textwidth]{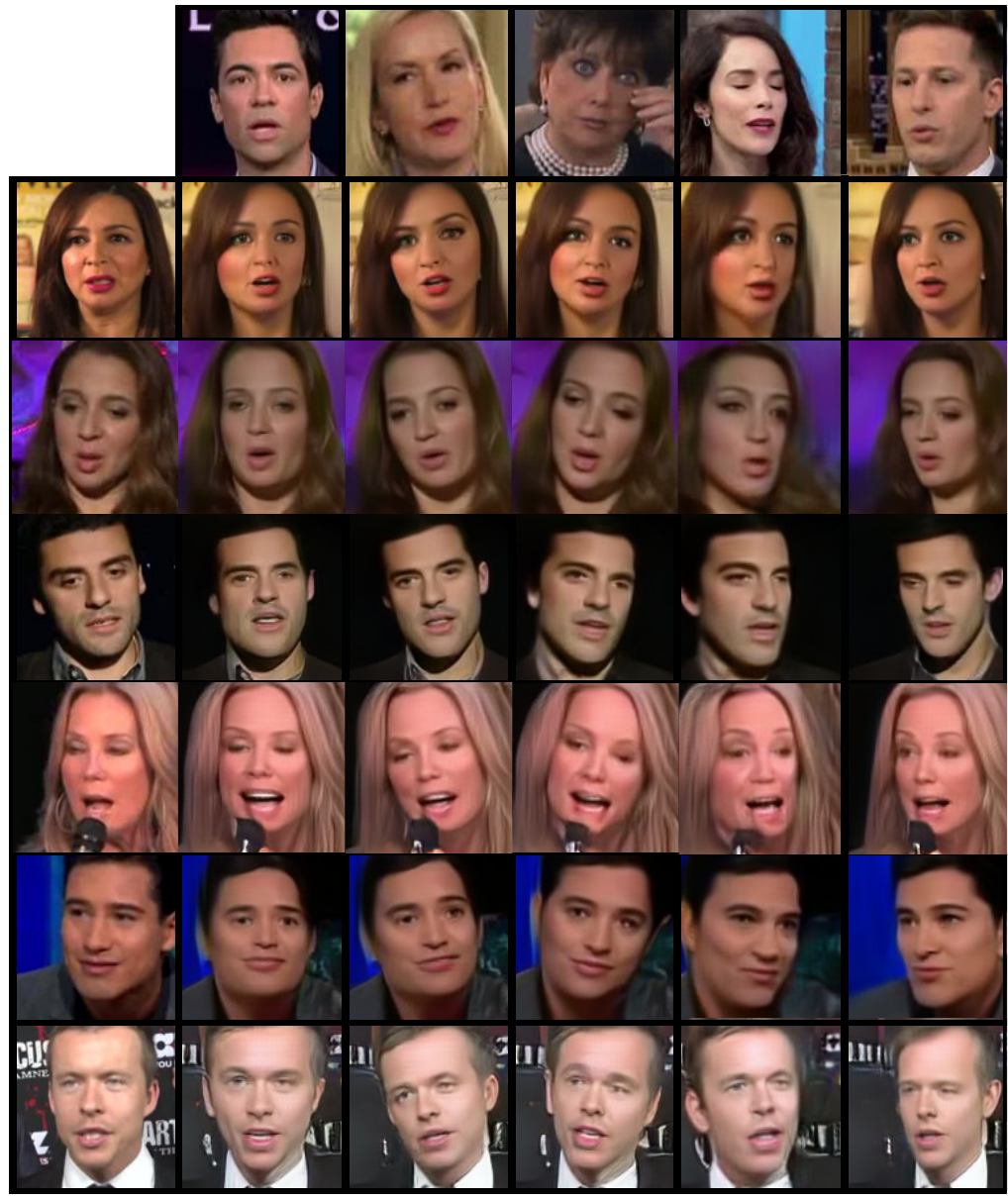}}
     \vspace{-0.8\baselineskip}    
    \subfloat[Mixed Reenactment] {\includegraphics[width=0.96\textwidth]{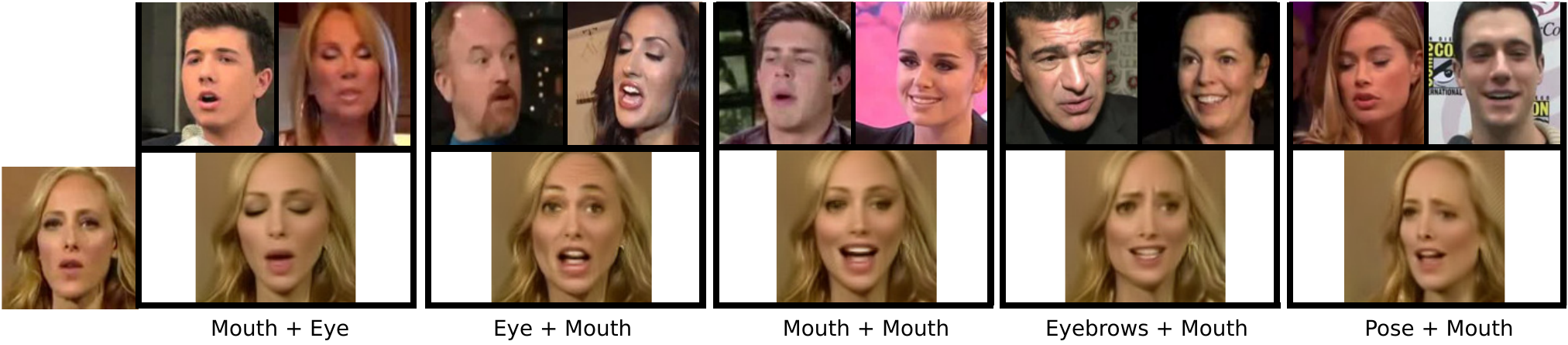}}
    \caption{Results for selective editing of facial attributes in face reenactment. (a-b) illustrate emotion and pose reenactment for various source images (extreme left column) and driving images (top row). (c) illustrates mixed reenactment by combining various attributes from source (extreme left) and two driving images (top row). The proposed method produces good quality results and provides control over the animation process, unlike other methods. More results are in the supplementary material. \vspace{-20pt}}
    \label{sedit}
\vspace{-0.2cm}   
\end{figure*}

\begin{figure*}
\begin{center}
\vspace{-5pt}
\includegraphics[width=12.5cm]{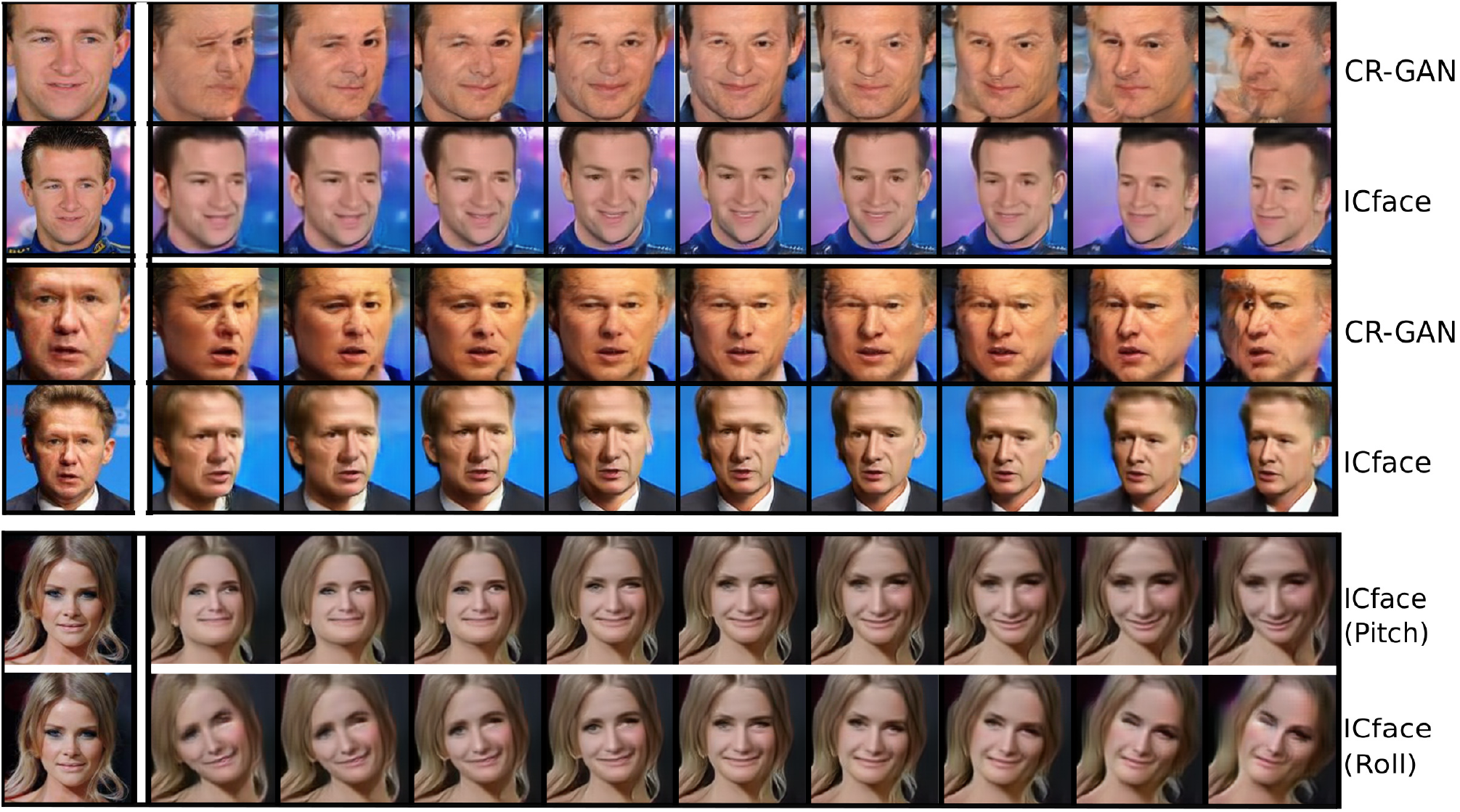}
\vspace{-3pt}  
\end{center}
   \caption{Results for multi-view face generation from a single view. In each block, the first row corresponds to CR-GAN \cite{c12} and the second row corresponds ICface. It is to be noted that each block contains the same identity with different crop sizes as both methods are trained with different image crops. Proposed architecture produces semantically consistent facial rotations by preserving the identity and expressions better than the CR-GAN \cite{c12}. The last two rows correspond to multi-view images generated from ICface by varying pitch and roll respectively which is not possible in CR-GAN \cite{c12}.\vspace{-0.3cm}}
\vspace{-0.3cm}   
\label{pose}
\end{figure*}
\subsection{Face Reenactment}
In face reenactment, the task is to generate a realistic face image depicting a given \emph{source} person with the same pose and expression as in a given \emph{driving} face image. The \emph{source} identity is usually specified by one or more example face images (one in our case). Figure \ref{x2face} illustrates several reenactment outputs using different \emph{source} and \emph{driving} images. We compare our results with two recent methods: X2Face \cite{x2face} and DAE \cite{c9}. We further refer to the supplementary material for additional reenactment examples. 

The results indicate that our model is able to retain the \emph{source} identity relatively well. Moreover, we see that the facial expression of the \emph{driving} face is reproduced with decent accuracy. Recall that our model transfers the pose and expression information via three angles and 17 AU activations. Our hypothesis was that these values are adequate at presenting the necessary facial attributes. This assumption is further supported by the result in Figure \ref{x2face}. Another important aspect in using pose angles and AUs, was the fact that they are independent of the identity. For this reason, the \emph{driving} identity is not ``leaking" to the output face (see also the results of the other experiments). Moreover, our model neutralises the \emph{source} image from its prior pose and expression which helps in reenacting new facial movements from \emph{driving} images. 
We assess this further in Section \ref{GANi}. 

\textit{Comparison to X2Face \cite{x2face}:} X2Face disentangles the identity (texture and shape of the face) and facial movements (expressions and head pose) using the \textbf{Embedding} and \textbf{Driving} networks, respectively. 
 It is trained unsupervisedly, which make it difficult to prevent all movement and identity leakages through the respective networks. These type of common artefacts are visible in some of the examples in Figure \ref{x2face}. We further note that the X2Face results are produced using three \emph{source} images unlike ICface with single source. 
 Additionally, the adversarial training of our system seems to lead to more vivid and sharp results than X2Face. To further validate this, we quantitatively compare the quality of generated images by both the methods. The quality is assessed in terms of image degradations like smoothing, motion blur, etc. by using two pre-trained networks, CNNIQA \cite{c40} and DeepIQA \cite{c41}, proposed for Non-Reference (NR) and Full-Reference (FR) Image Quality Assessment (IQA). For the FR-IQA, the source image is used as the reference image. The mean quality scores over all the test images for X2face and ICface are presented in Table \ref{TQ}. The lower scores for ICface signify that the reenacted faces are less distorted than X2Face in comaprision with an ideal imaging model or given reference image. 

\begin{table}
\begin{center}
\begin{tabular}{|l|c|c|}\hline
\small Method & \small CNNIQA(NR) & \small DeepIQA (FR) \\
\hline\hline
\small X2face / ICface & \small 28.89 / \textbf{25.02} & \small 39.49 / \textbf{33.08} \\
\hline
\end{tabular}
\end{center}
\caption{Image Quality Assesment scores (Lower is better.)\vspace{-0.5cm}}
\vspace{-0.5cm}
\label{TQ}
\end{table}

\textit{Comparison to DAE \cite{c9}:} DAE proposed a special autoencoder architecture to disentangle the appearance and facial movements into texture image and deformation fields. We trained their model on VoxCeleb dataset \cite{c23} using the publicly available codes from the original authors. For reenactment, we first decomposed both the \emph{source} and \emph{driving} images into corresponding appearances and deformations. Then we reconstructed the output face using the appearance of \emph{source} image and the deformation of \emph{driving} image. The obtained results are presented in Figure \ref{x2face}.
The DAE often fails to transfer the head poses and identity accurately. The head pose related artefact is best observed when the pose difference between the \emph{source} and \emph{driving} is large. These challenges might be related to the fact that the deformation field is not free from the identity specific characteristics. 

\vspace{-0.2cm}
\subsection{Controllable Face Reenactment} 

The pure face reenactment animates the \emph{source} face by copying the pose and expression from the \emph{driving} face. In practice, it might be difficult to obtain any single \emph{driving} image that contain all the desired attributes. The challenge is further emphasised if one aims at generating an animated video. Moreover, even if one could record the proper \emph{driving} frames, one may still wish to perform post-production editing on the result. This type of editing is  hard to implement with previous methods like X2Face \cite{x2face} and DAE \cite{c9} since the facial representation is learned implicitly and it lacks a clear interpretability. Instead, the head pose angles and AUs, utilised in our approach, provide human interpretable and easy-to-use interface for selective editing. Moreover, this presentation allows to mix attributes from different \emph{driving} images in controlled way. Figures \ref{teaser} and \ref{sedit} illustrate multiple examples, where we have mixed the \emph{driving} information from different sources. The supplementary material contains further example cases.

\begin{figure}[t]
\begin{center}
\vspace{-0.15cm}
   \includegraphics[width=8.1cm]{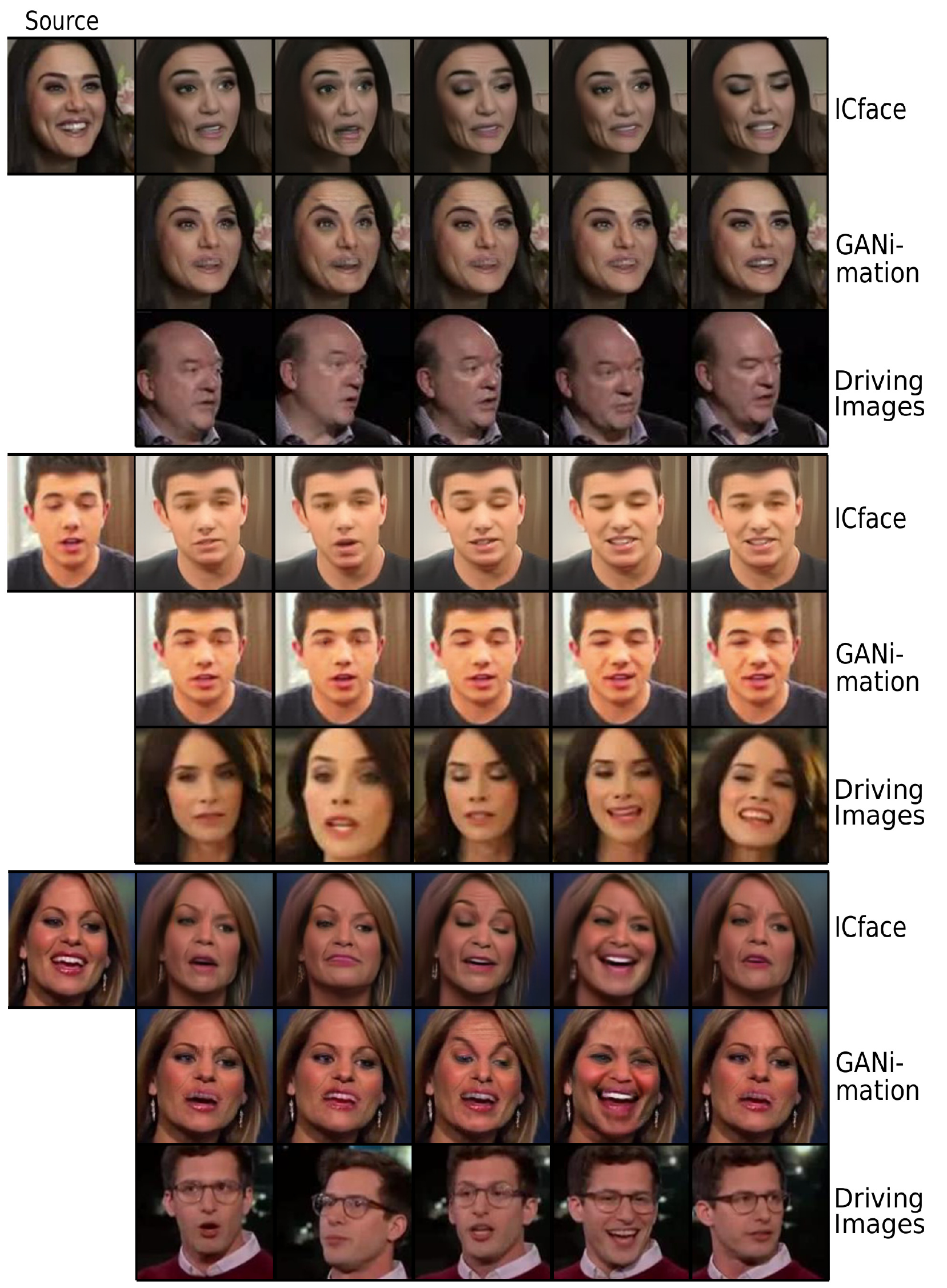}
\end{center}
\vspace*{-10pt}
   \caption{The results for manipulating emotion in the face images. For each source image, the first row is generated using ICface, the second row using GANimation \cite{gani} and the third row contains the driving images. As ICface first neutralises the source image, it is evident that it produces better emotion reenactment when the source has initial expressions (first and third row).\vspace{-0.6cm}}
\vspace{-0.3cm}
\label{GANI}
\end{figure}

\begin{table}
\begin{center}
\begin{tabular}{|p{15mm}|p{8mm}|p{8mm}|p{8mm}|p{1cm}|p{8mm}|}\hline 
\small Measures & \small X2face & \small GANi-mation & \small DAE & \small ICface \scriptsize (No-Neutralizer) & \small ICface  \\
\hline
\small Accuracy\scriptsize(\%) & \small 64.76 & \small 61.77 & \small 59.86 & \small 59.88 & \small 62.86 \\
\small F-score & \small 0.4476 & \small 0.3944 & \small 0.3734 & \small  0.3759 & \small 0.4185 \\
\hline
\end{tabular}
\end{center}
\vspace{0.5cm}
\caption{Comparision of Action units classification measures (Higher is better)\vspace{-0.5cm}}
\vspace{-0.5cm}
\label{TAU}
\vspace{-0.5cm}
\end{table}

\subsection{Facial expression Manipulation} \label{GANi}

In this experiment, we concentrate on assessing how the proposed model can be used to transfer only the expression from the \emph{driving} face to the \emph{source} face. We compare our results to GANimation \cite{gani} that is a purpose-built method for manipulating only the facial expression (i.e.~it is not able to modify the pose). 
Figure \ref{GANI} illustrates example results for the proposed ICface and the GANimation. Note that in Figure \ref {GANI} the head pose of ICface is kept constant as that of the source image only for better illustration. The latter method seems to have challenges when the \emph{source} face has an intense prior expression that is different from the \emph{driving} expression. In contrast, our model neutralised the \emph{source} face before applying the \emph{driving} attributes. This approach seems to lead in better performance in the cases where \emph{source} has intense expression. We have further validated this qualitatively by calculating the facial action consistency. This is obtained by comparing the presence and absence of
AUs in both the driving and the reenacted images. We used a pre-trained network
\cite{c28} to predict the presence of 17 action units in a yes or
no fashion for the images generated by X2face, DAE, ICface and
GANimation. Then we calculated the balanced accuracy \cite{c39} and F-score \cite{c39} of detecting the presence and absence of AUs in the generated images in comparison to those of the driver images. These measures are chosen as the absent AU counts are significantly more than active AUs. The values are listed in Table \ref{TAU} and ICface achieves higher or comparable scores than others even after using only 17 parameters for representing all the facial activities. 

\subsection{Multiview Face generation}

Another interesting aspect in face manipulation is the ability to change the viewpoint of a given \emph{source} face. Previous works have studied this as an independent problem. We compare the performance of our model in this task with respect to the recently proposed Complete-Representation GAN (CR-GAN) \cite{c12} method. The CR-GAN is a purpose-built method for producing multi-view face images from a single view input. The model is further restricted to consider only the yaw angle of the face. The results in Figure \ref{pose} were obtained using the CR-GAN implementation from the original authors \cite{c30}. Their implementation was trained on CelebA \cite{c30} dataset, and therefore we used CelebA \cite{c30} test to produce these examples. We note that we did not re-train or fine-tune our model on CelebA. The results indicate that our model is able to perform facial rotation with relatively small semantic distortions. Moreover, last two rows of Figure \ref{pose} depict rotation along pitch and roll axis which is not achievable with the CR-GAN. We believe that our two-stage based approach \((input \rightarrow neutral \rightarrow target)\) is well suited for this type of rotation tasks.

\begin{figure}[t]
\begin{center}
   \vspace{-0.2cm}
   \includegraphics[width=8.0cm]{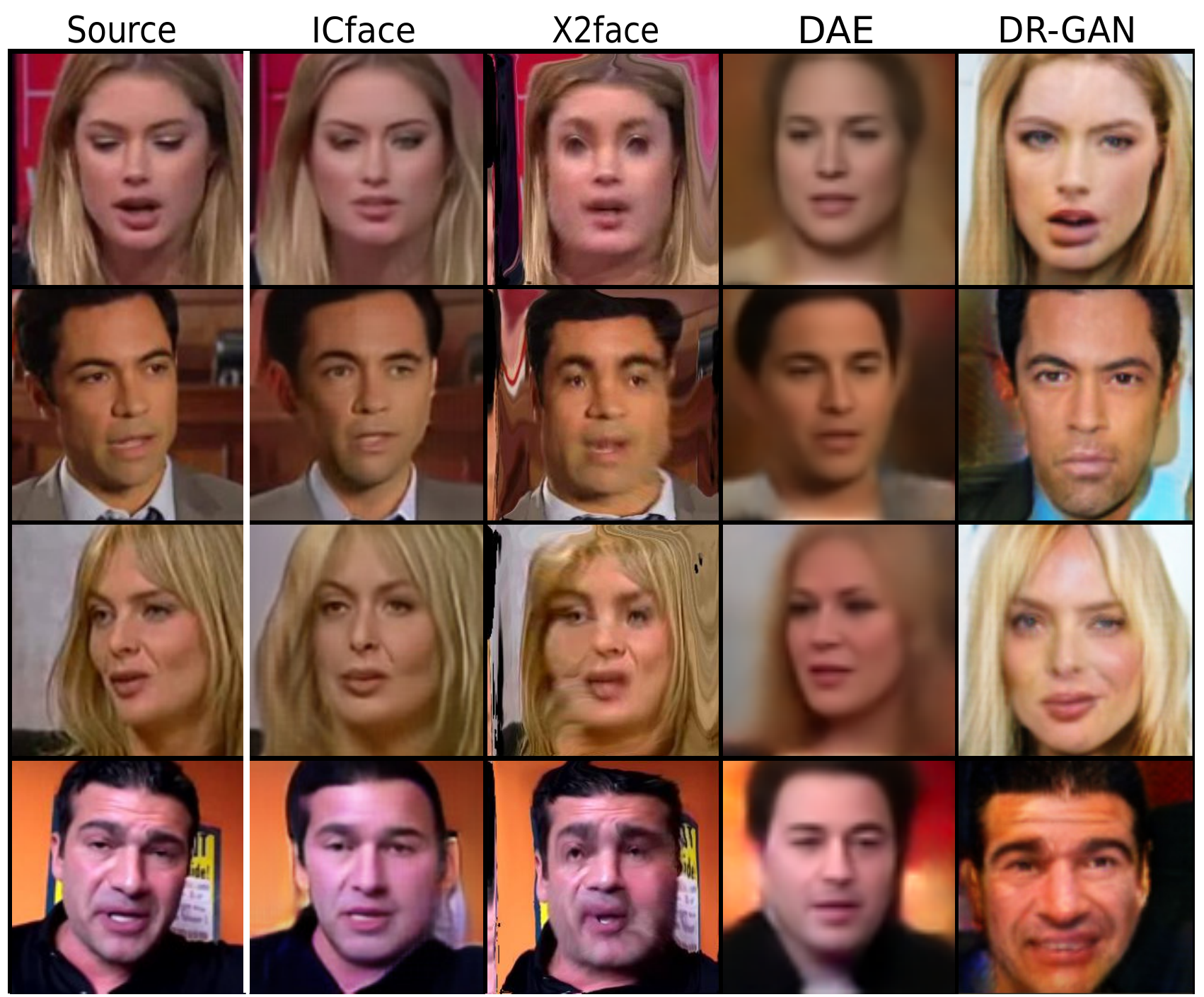}
\end{center}
\vspace*{-8pt}
   \caption{The results for generating neutral face from a single source image. The proposed method produces good image quality even with extreme head poses (third row).\vspace{-0.5cm}}
\vspace{-0.1cm}
\label{neut}
\end{figure}
\subsection{Identity disentanglement from face attributes}
Finally, in Figure \ref{neut}, we demonstrate the performance of our neutraliser network. The neutraliser was trained to produce a template face from the single \emph{source} image with frontal pose and no expression. We believe that the effective neutralisation of the input face is one of the key reasons why our system produces high quality results in multiple tasks. To verify that, We repeated the experiments by removing neutralization part from our model and the performance decreases as given in Table \ref{TAU} (Fourth column). Figure \ref{neut} also illustrates the neutral images (or texture image) produced by the baseline methods \footnote{It is not a direct comparision to X2face \cite{x2face} as it never aims at generating a neutral image. It only illustrates the intermediate results for X2face.}. One of these is DR-GAN \cite{c34} that is a purpose-built face frontalisation method (i.e.~it does not change the expression). The ICface successfully neutralises the face while keeping the identity intact even if the \emph{source} has extreme pose and expression.   

\vspace{-0.2cm}
\section{Conclusion}

In this paper, we proposed a generic face animator that is able to control the pose and expression of a given face image. The animation was controlled using human interpretable attributes consisting of head pose angles and action unit activations. The selected attributes enabled selective manual editing as well as mixing the control signal from several different sources (e.g. multiple \emph{driving} frames). One of the key ideas in our approach was to transform the \emph{source} face into a canonical presentation that acts as a template for the subsequent animation steps. Our model was demonstrated in numerous face animation tasks including face reenactment, selective expression manipulation, 3D face rotation, and face frontalisation. 
In the experiments, the proposed ICface model was able to produce high quality results for a variety of different \emph{source} and \emph{driving} identities. The future work includes further increasing the resolution of the output images and further improving the performance with extreme poses having a few training samples

{\small
\bibliographystyle{ieee}
\bibliography{egbib}
}

\end{document}